\documentclass[conference]{IEEEtran}
\IEEEoverridecommandlockouts
\usepackage{cite}
\usepackage{amsmath,amssymb,amsfonts}
\usepackage{algorithmic}
\usepackage{graphicx}
\usepackage{textcomp}
\usepackage{xcolor}
\usepackage{graphicx}
\usepackage{ctable} 
\usepackage{multicol}
\usepackage{multirow}
\usepackage{amsfonts}
\usepackage{caption}

\def\BibTeX{{\rm B\kern-.05em{\sc i\kern-.025em b}\kern-.08em
    T\kern-.1667em\lower.7ex\hbox{E}\kern-.125emX}}

\begin{document}

\title{Initial Test of ``BabyRobot'' Behaviour on a Teleoperated Toy Substitution: Improving the Motor Skills of Toddlers}

\author{\IEEEauthorblockN{Eric Canas}
\IEEEauthorblockA{\textit{FIB} \\
\textit{UPC BarcelonaTech}\\
Barcelona, Spain \\
eric.canas@estudiantat.upc.edu}
\and
\IEEEauthorblockN{Alba M. Garcia}
\IEEEauthorblockA{\textit{FIB} \\
\textit{UPC BarcelonaTech}\\
Barcelona, Spain \\
alba.maria.garcia@estudiantat.upc.edu}
\and
\IEEEauthorblockN{Anais Garrell}
\IEEEauthorblockA{\textit{IRII (CSIC-UPC)} \\
\textit{UPC BarcelonaTech}\\
Barcelona, Spain \\
agarrell@iri.upc.edu}
\and
\IEEEauthorblockN{Cecilio Angulo}
\IEEEauthorblockA{\textit{IRII (CSIC-UPC), IDEAI-UPC} \\
\textit{UPC BarcelonaTech}\\
Barcelona, Spain \\
cangulo@iri.upc.edu}
}

\maketitle

\thispagestyle{empty}
\pagestyle{empty}

\begin{abstract}
This article introduces ``Baby Robot'', a robot designed to improve infants' and toddlers' motor skills. This robot is a car-like toy that moves autonomously by using reinforcement learning and computer vision. Its behaviour consists of escaping from a target infant that has been previously recognized, or at least detected, without compromising the infant's security by avoiding obstacles. Regarding other robots that share this purpose, there is a variety of commercial toys available on the market; however, no one is betting on an intelligent autonomous movement, since they use to repeat simple, yet repetitive movements. In order to examine how that autonomous movement may improve infants' mobility, two crawling toys --one in representation of ``Baby Robot''-- were tested in a real environment. These real-life experiments were conducted with a safe and approved surrogate of our proposed robot in a kindergarten, where a group of infants interacted with the toys. Improvements in the efficiency of the play-session were detected.
\end{abstract}

\begin{IEEEkeywords}
crawling, infant, toddler, toy, robot, motor skills
\end{IEEEkeywords}

\section{INTRODUCTION}
Crawling is a key step in the locomotion evolution for most infants, which concludes when the infant is able to walk. For approximately 50\% of them~\cite{bib:main-reference-crawling}, crawling behaviour usually starts at the age of 8 months, but it can also be later or never happening. It is an issue of interest for paediatric professionals and parents, as it is a common concern about children motor development. The current state of the art about the development of crawling in infants enumerate three key factors that ease its appearance:
\begin{itemize}
    \item Smaller, slimmer, more maturely proportioned infants tend to crawl at earlier ages than larger, chubbier infants~\cite{bib:locomotion}. Hence, infants with a favourable ratio of muscle to body fat have a clear advantage in mobility.
    \item Infants that spend more time in prone position when they are awake tend to crawl earlier~\cite{bib:tummy-time}. These sessions, commonly known as ``tummy time'', help them to strengthen and control better key muscles for crawling, such as the ones in their neck and shoulders, among others \cite{bib:tummy2}. Additionally, early promotion of ``tummy time'' has been shown to be effective in improving feeding practices on infants between 1 and 12 months \cite{bib:feed}, as well as reducing the motor delay on infants with \textit{Down syndrome}~\cite{bib:down}.
    \item And, finally, positioning a certain toy out of the reach of infants encourages them to move towards it~\cite{bib:toy-out-of-reach}~\cite{bib:journey}. When this process is iterated several times, that is, the toy position is changed again when the infant is getting closer to it, longer training sessions may be obtained. Additionally, it is important to let them play with the toy after several trials in order to avoid frustration.
\end{itemize}

\begin{figure}
    \centering
    \includegraphics[width=0.8\linewidth]{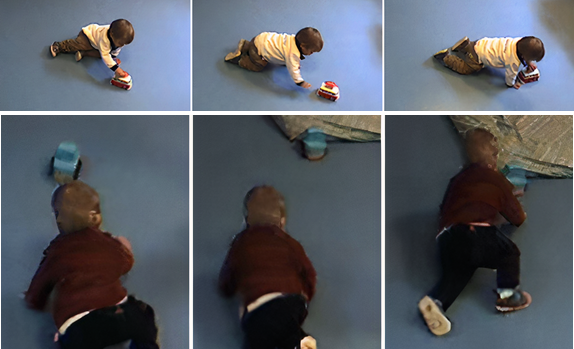}
    \caption{Toddlers from a kindergarten interacting with the ``crawling toy'' used for the control condition~(top) and the ``Baby Robot'' condition~(bottom).}\label{fig:intro}
    \end{figure}

The last point has encouraged paediatric researchers and toy manufacturers to develop numerous ``crawling toys'', assistive robots~\cite{bib:sippc}~\cite{bib:interface} and even whole environments \cite{bib:gear}.

In this study, we aim to investigate how the autonomous movement of ``crawling toys'' --toys designed to help parents and caregivers to develop children's motor skills by encouraging them to crawl independently \cite{bib:sar} (see Fig.~\ref{fig:intro})-- affects infants' engagement in crawling. Our main hypothesis is that, when compared to toys that do not consider their sensory feedback, fully autonomous and intelligent toys improve infants' engagement with crawling activities. That is, we advocate for physical agents not implementing iterative movements, but robot motion behaviour taking children sensory feedback into consideration. For this reason, we introduce ``Baby Robot'', a crawling robot designed to improve the mobility of infants and toddlers. ``Baby Robot'' is a car-like toy that moves autonomously using reinforcement learning and computer vision. It escapes from an infant that has been recognized, or at least detected, while avoiding obstacles, so that the security of the infant is not compromised. 

The rest of the paper is organized as follows: First, we will introduce ``Baby Robot'', describing its main purpose, behaviour and skills. Once the context of ``Baby Robot'' is defined, we will describe how the experiments were conducted, in order to accept or refuse our hypothesis. Hence, in Section \ref{sec:methodology} we introduce the methodology of the current work. Section \ref{sec:evaluation} presents the evaluation and the obtained results from the experimentation. Finally, Section \ref{sec:results} and \ref{sec:discussion} describe the results and the discussion about the results yielded, respectively.

\section{Methodology} \label{sec:methodology}
This section presents our research, its purpose, the target actors, and the development of the robot's behaviour.

\subsection{Purpose}
``Baby Robot's" purpose is the same as any other ``crawling toy", which is to encourage infants to crawl --both those who are still learning and those who have already begun-- by placing itself out of the infant's reach in order to catch their attention and motivate them to chase it. 

\subsection{Target actors}
``Baby Robot'' aims to be a ``crawling toy'', so it is expected to be used in scenarios where the upbringing of infants at the earlier stages happen, such as their homes or kindergarten. It is recommended that these spaces comply with some characteristics to make the crawling sessions safer and better: they should be closed in order to control better the behaviour of the infant so as to avoid any possible risk. Additionally, they should also be clear, in order to provide the infant with a large enough space to play. To sum up, we expect that the responsible for the infant places ``Baby Robot" in a place where it can move freely without major concerns for their safety and under the supervision of the aforementioned responsible(s). Therefore, we can identify three clear roles that appear during this interaction:
\begin{itemize}
    \item \textbf{``Baby Robot''}, whose purpose is to motivate the infant to crawl by following it.
    \item The \textbf{infant}, who gets interested in the robot and tries to chase it.
    \item The \textbf{adult responsible of the infant} at the time, who supervises the interaction and intervene when the infant is at risk or needs help, when the robot is not functioning as expected or when they switch the robot off to let the infant play with it or to end the session.
\end{itemize}

\subsection{Behaviour of the robot}\label{sec:behaviour}

\begin{figure}
    \centering
    \includegraphics[width=0.95\linewidth]{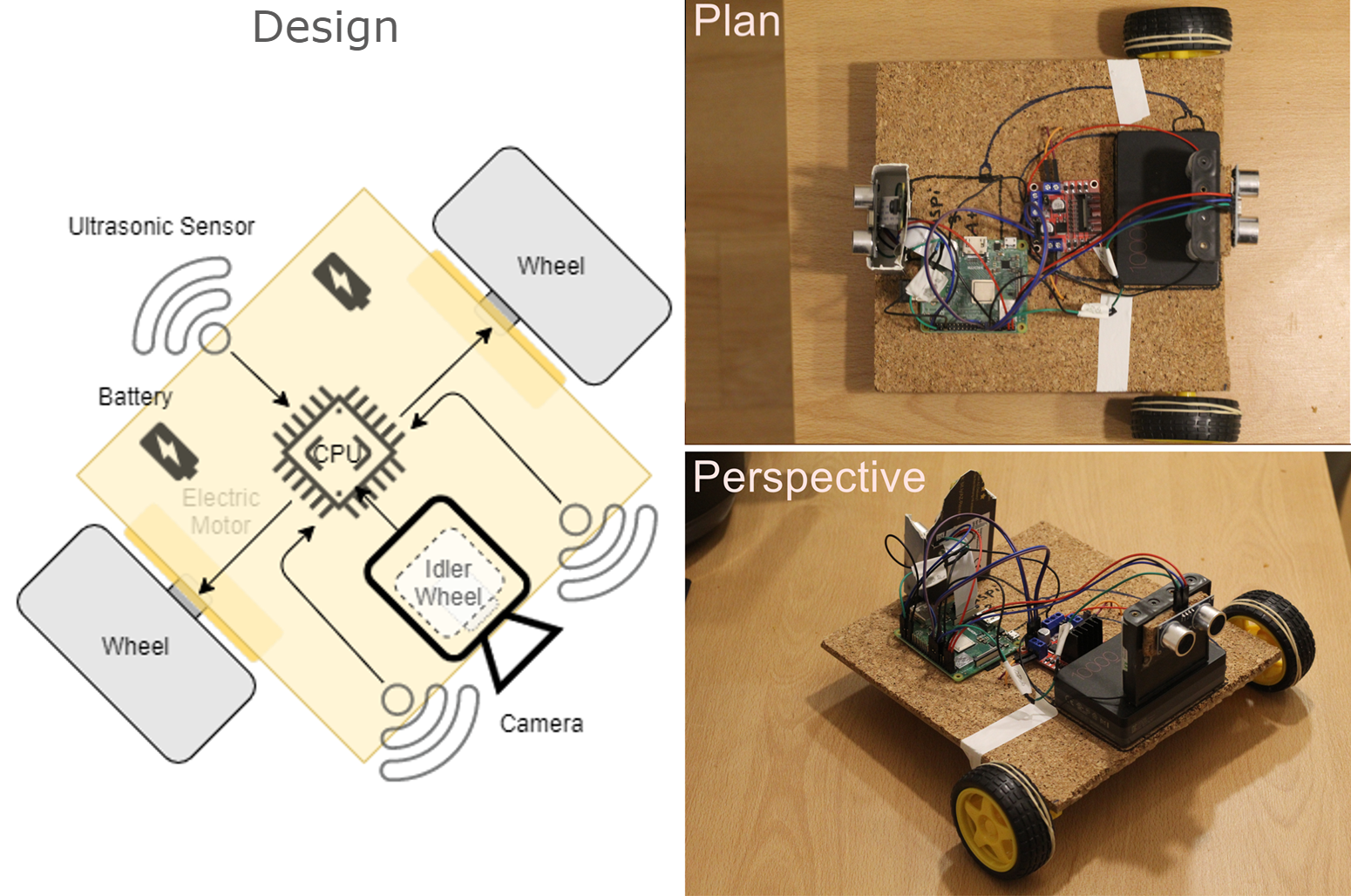}
    \caption{Simplified scheme of the ``Baby Robot'' structure (left) and view of the initial prototype (right).}
    \label{fig:robotScheme}
\end{figure}

The ultimate goal of the behaviour of ``Baby Robot'' is to escape from the target infant, who is expected to follow the robot, while maintaining a constant distance of separation with them and avoiding any possible obstacle. This goal is achieved by means of the movement that allow the two wheels --with their corresponding electrical motors-- and the idler wheel the robot has equipped along with the camera and the ultrasonic sensors placed at the front and at the back of the robot, respectively. A simplified scheme of the robot structure can be observed in Fig.~\ref{fig:robotScheme}.

\begin{figure}
    \centering
    \includegraphics[width=0.95
    \linewidth]{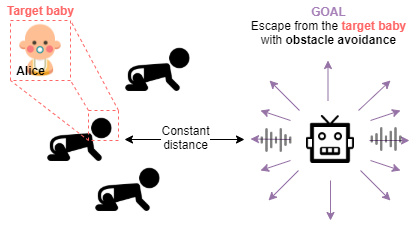}
    \caption{Diagram of the ``Baby Robot'' behaviour.}
    \label{fig:locomotion}
\end{figure}

``Baby Robot'' can move in a total amount of twelve directions: forward, backwards, left, right, the four classic diagonals that are separated from the previous directions by a 45$^{\circ}$ angle and four extra orientations that are close to the forward and backwards directions forming a little angle to their left and their right. We added the latter as this little angle could help it to achieve softer trajectories. Of course, it can also decide not to move, as well as to make different movements in place to catch the attention of the toddler when it is temporally lost. A diagram of the ``Baby Robot'' behaviour is shown in Fig.~\ref{fig:locomotion}.

The decision about which movement should be executed at each time-step is decided exclusively according to the distance and horizontal deviation to the target infant and distance to the closer obstacle. When the infant is not detected, ``Baby Robot'' uses the prior information about its latest position in order to rapidly locate it again.

\subsection{Cognitive skills of the robot}

``Baby Robot'' interacts with their target users by means of its movement exclusively. Its objective is to escape from an infant that has been identified at the beginning of the session as its target. The way our robot is able to decide which action must perform comes from two different sources that summarize its cognitive skills:
\begin{itemize}
    \item The ability to recognize human bodies and faces, to identify faces and to compute the distance and horizontal deviation to these detections by means of the camera placed in front of the robot.
    \item The ability to compute the distance to the closest obstacles by means of its front and back ultrasonic sensors.
\end{itemize}

\section{Evaluation} \label{sec:evaluation}
This section is devoted to the experimentation performed over the implemented robot, ``Baby Robot''. First, we specify all the test performed as well as the data obtained from them. Next, we analyse these data in order to accept or reject our initial hypothesis: \textit{When compared to toys that don't consider their sensory feedback, fully autonomous and intelligent toys improve infants' engagement with crawling activities}.

\subsection{Toys implied}
\begin{figure}
    \centering
    \includegraphics[width=0.95\linewidth]{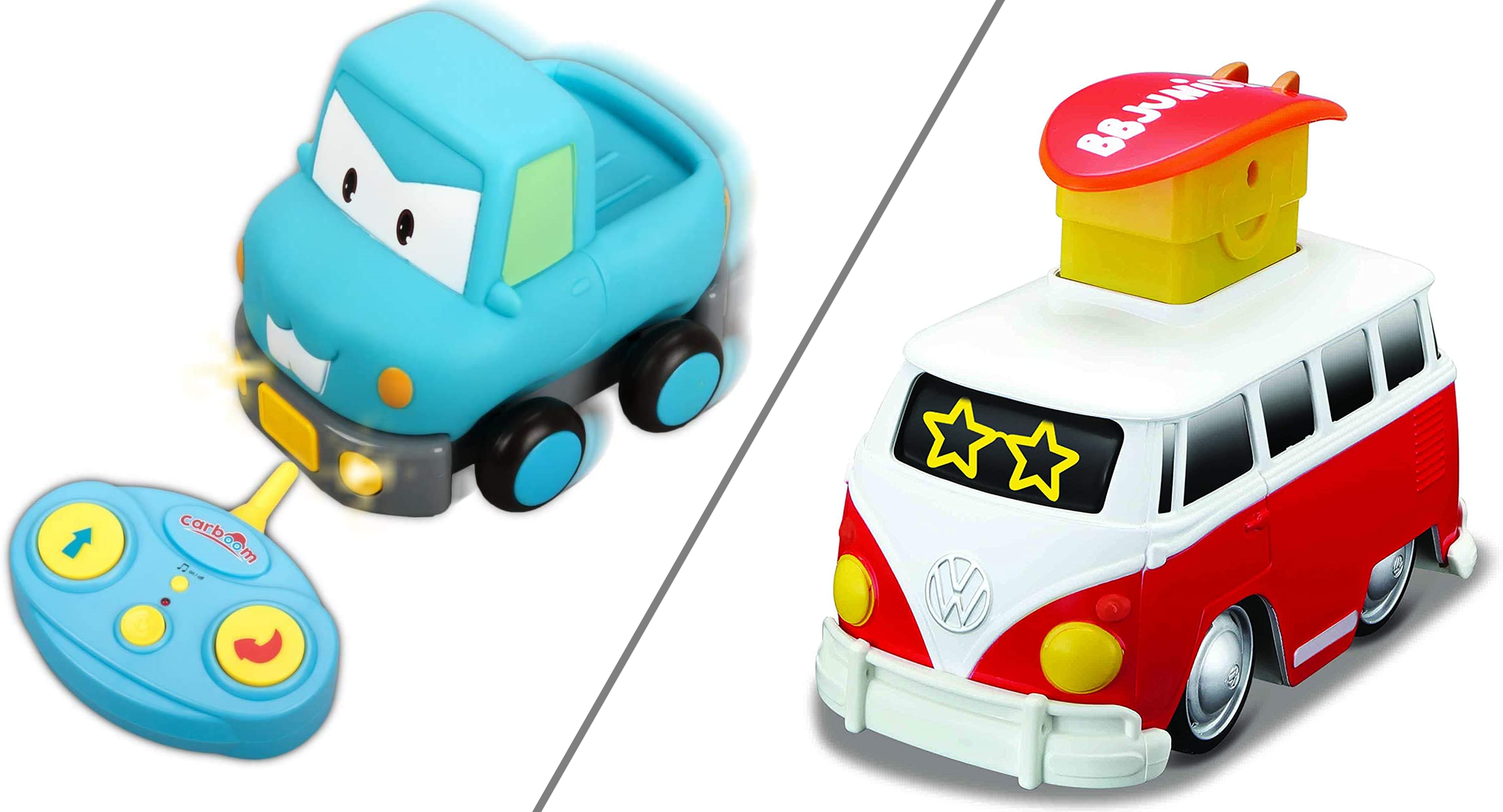}
    \caption{Teleoperated toy in substitution of ``Baby Robot'' (left) and \textit{press \& go} toy selected for the control condition (right).}\label{fig:bothToys}
\end{figure}

As ``Baby Robot'' is a prototype, it would be unsafe for infants to carry out the experimentation with it due to its accessible wires and batteries. Therefore, we have substituted it with a car-like toy by ToyTown that can be teleoperated in order to simulate ``Baby Robot'' behaviour (see Fig.~\ref{fig:bothToys}~(left)). 
For the control condition, we have used a car-like press-and-go toy by TaviToys (see Fig.~\ref{fig:bothToys}~(right)). This toy has been selected because it is, functionally and aesthetically, the closest option to the one used for the ``Baby Robot'' condition, but without the autonomous and intelligent movement capabilities. It is important to remark that both toys have been selected aiming to isolate the ``movement variable'', thus avoiding the influence of lights, sounds or aesthetics during the experimentation.

\subsection{Test}\label{sec:tests}

Our experimentation was performed over a limited set of 6 toddlers from a kindergarten, aged between 7 and 16 months. Their abilities ranged from those just taking their first crawling steps up to those who were already able to walk, but that still crawled frequently. We selected half of these toddlers for the control condition and the other half for the ``Baby Robot'' condition, separating --with the help of their caregivers --each one of the two most skilled and two least skilled infants in different groups, thus maximizing the homogeneity of the groups. In both conditions, they played with their respective toys during five sessions of ten minutes, distributed along consecutive days. All the sessions were performed in a familiar environment for the toddlers, that were always accompanied by their caregiver --an adult of their trust--. In the control condition, the caregiver showed them how the \textit{press \& go} toy worked, and let the infant continue playing with it. For the ``Baby Robot'' condition, the caregiver put the toy near to the infant and secretly used the remote control for simulating that the car autonomously escaped from the toddler. For both conditions we measured the following variables:
\begin{enumerate}
    \item Percentage of time that the toddler is in movement during the session.
    \item Distance travelled by the toddler during the session.
\end{enumerate}

We must take into consideration the following clarifications:
\begin{itemize}
    \item To prevent our presence from causing any disturbance in the toddler behaviour, the unique other person inside the room was always their usual --and trusted-- caregiver. We always observed the sessions through a webcam placed in the corner of the room.
    
    \item Time that the toddler is rotating over itself, is considered as movement time.
    
    \item Distances travelled are estimated from a set of discrete landmarks placed in the room and distanced at one meter each one. Nevertheless, trajectory measurements could present approximation errors.
\end{itemize}

\begin{figure}
    \centering
    \includegraphics[width=0.95\linewidth]{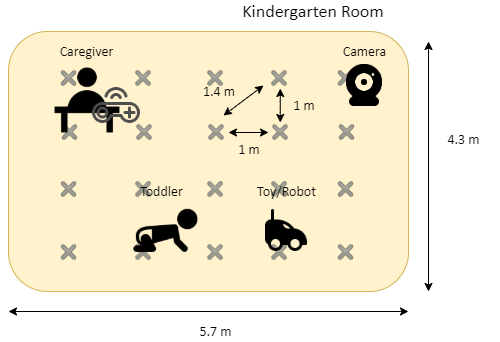}
    \caption{Experimental environment setup.}\label{fig:environment}
\end{figure}

A simplified scheme of the proposed experimental setup is presented in Fig.~\ref{fig:environment}.

\section{Results} \label{sec:results}
\begin{figure}
    \centering
    \includegraphics[width=0.95\linewidth]{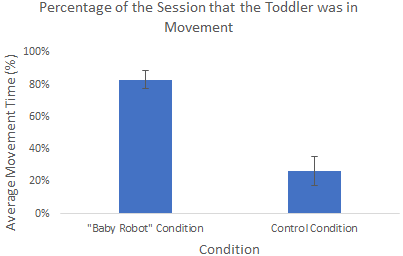}
    \caption{Percentage of the session that the toddlers were in movement.}\label{fig:avgMovementTime}
\end{figure}

Fig.~\ref{fig:avgMovementTime} represents the percentage of the session that the toddler was in movement according to both conditions: ``Baby Robot'' condition and control condition. There is an observable tendency depicted in Fig.~\ref{fig:avgMovementTime}: In ``Baby Robot'' sessions, toddlers were in movement during the $82.26\pm5.6\%$ of the session, while toddlers of the control condition were only in movement during the $26.55\pm8.9\%$ of it. Therefore, in terms of training, ``Baby Robot'' produced a $3.1\times$ improvement in the efficiency of the session. 

\begin{figure}
    \centering
    \includegraphics[width=0.95\linewidth]{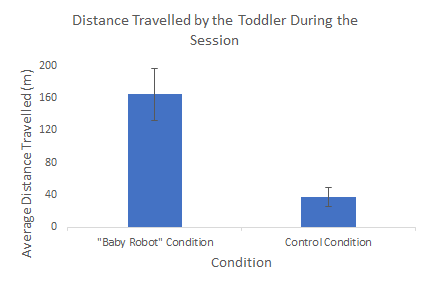}
    \caption{Average distance travelled by toddlers in each session.}\label{fig:avgDistance}
    \end{figure}

From the point of view of the distance travelled by the toddler during the session, Fig.~\ref{fig:avgDistance}, also describes a tendency: In ``Baby Robot'' sessions, infants travelled in average $164.37\pm31.7$ meters, while infants in the control condition travelled $37.6\pm11.9$ meters in the same amount of time. This enhances the efficiency of the play session in a $4.4\times$ factor.

In qualitative terms, caregivers in charge of controlling both toys reported the following key conclusions:
\begin{itemize}
    \item Toddlers were clearly more engaged with the activity in the ``Baby Robot'' condition.
    \item They moved faster and made longer trajectories in the ``Baby Robot'' condition.
    \item Although the disengagement process was slowed down in ``Baby Robot'' condition, in both cases the engagement with the activity decreased as the session progressed. 
\end{itemize}


Although we acknowledge that the limited sample size of this initial test limits the statistical relevance of these results, we ran a \textit{t-test} over them, obtaining $p<0.0001$. For this reason, and with these preliminary data, we are encouraged to reject the null hypothesis, accepting that: \textit{Infants may be more engaged to crawl when the toy presents intelligent and autonomous movements, rather than when their movements are neither intelligent nor autonomous}.

\section{Discussion} \label{sec:discussion}
In this study, we investigated how toys with intelligent and autonomous movements that take into account the sensory feedback of the infants affect their engagement with the crawling activity. We found preliminary support to our initial hypothesis. This contributes to the open line of research that aims to improve the engagement, and therefore the efficiency, of children in their first steps of apprenticeship. We hope that further research on these \textit{toys/robots} will help parents around the world to promote the early and healthy development of their children movement abilities, providing them with all the benefits that these practices present for their correct growth.

\subsection{Limitations}
Even though the conclusions described in this initial test support our initial hypothesis clearly, we acknowledge that they should be replicated in a wider population and that they could be slightly noised by some uncontrollable factors that limited the isolation of the ``autonomous/intelligent movement'' variable as well. These factors are, for example, the lack of infants with special mobility conditions or any difference in the caregiver's behaviour between both activities that may enhance the engagement of some infant in one of the testing conditions. Despite these considerations, we believe that these results are a good starting point and that the validity of the proposed hypothesis should be kept in a wide range of cases.

\subsection{Future work}
 
Future research in this field shall continue to investigate how sounds, lights, colours and aesthetics affect the engagement that this kind of robots produce in children. Although this is one of the key points that the industry has faced when developing toys, there are no studies on how these factors affect to the stimulation of the toddler when playing with ``crawling toys''. We consider that these lights, sounds, colours and cute aesthetics would play a key role in the study of how to keep the attention of the toddler for longer periods of time, thus increasing even more the effectiveness of the training sessions. Additionally, it would be necessary to expand the generality of the conclusions extracted from this study by researching in more diverse groups, including children with very diverse cultural backgrounds and different mobility conditions, as well as toddlers in all the stages of their early childhood. Finally, it should be studied how toys like ``Baby Robot'' could accelerate the transitions between different mobility stages, it is, the transition from ``tummy time'' to the crawling stage or from crawling to walking.
\addtolength{\textheight}{-15.288cm}   




\vfill\eject

\end{document}